\documentclass[10pt,twocolumn,letterpaper]{article}

\usepackage{iccv}
\usepackage{times}
\usepackage{epsfig}
\usepackage{graphicx}
\usepackage{amsmath}
\usepackage{amssymb}

\usepackage{booktabs}
\usepackage{enumerate}
\usepackage{makecell}

\usepackage[normalem]{ulem}
\usepackage[para]{footmisc}
\usepackage{color}

\usepackage{wrapfig}

\usepackage{subcaption}
\usepackage{multirow}
\usepackage{eqparbox,array}
\usepackage{enumerate}

\usepackage{isotope}
\usepackage{siunitx}
\usepackage{tablists}
\usepackage{bm}
\usepackage[table]{xcolor}
\usepackage{ulem}

\usepackage{algorithm}
\usepackage{listings}
\usepackage{enumitem}
\usepackage{pifont}

\usepackage{etoolbox}
\makeatletter
\AfterEndEnvironment{algorithm}{\let\@algcomment\relax}
\AtEndEnvironment{algorithm*}{\kern2pt\hrule\relax\vskip3pt\@algcomment}
\let\@algcomment\relax
\newcommand\algcomment[1]{\def\@algcomment{\footnotesize#1}}
\renewcommand\fs@ruled{\def\@fs@cfont{\bfseries}\let\@fs@capt\floatc@ruled
  \def\@fs@pre{\hrule height.8pt depth0pt \kern2pt}%
  \def\@fs@post{}%
  \def\@fs@mid{\kern2pt\hrule\kern2pt}%
  \let\@fs@iftopcapt\iftrue}
\makeatother

\newcommand{\ourmethod}{CTP}
\newcommand{\ourdataset}{P9D}

\newcommand{\cmark}{\ding{51}}
\newcommand{\xmark}{\ding{55}}

\newcommand{\fig}{{Figure}\@\xspace}
\newcommand{\tab}{{Table}\@\xspace}

\newcommand{\alg}{{Alg.}\@\xspace}

\usepackage[pagebackref=true,breaklinks=true,letterpaper=true,colorlinks,bookmarks=false]{hyperref} 

\usepackage[capitalize]{cleveref}
\crefname{section}{Sec.}{Secs.}
\Crefname{section}{Section}{Sections}
\Crefname{table}{Table}{Tables}
\crefname{table}{Tab.}{Tabs.}

\definecolor{darkgreen}{rgb}{0,0.6,0.2}
\definecolor{danred}{rgb}{0.9098,0.9098,0.9098}
\definecolor{shenred}{rgb}{0.8117,0.8117,0.8117}

\iccvfinalcopy 

\def\iccvPaperID{***} 

\begin{document}

\title{
\vspace{-8mm}
{\ourmethod}: Towards Vision-Language Continual Pretraining via Compatible Momentum Contrast and Topology Preservation}
\vspace{-8mm}
\author{
Hongguang Zhu$^{1,2}$\footnotemark[1]\quad Yunchao Wei$^{1,2,3}$\quad Xiaodan Liang$^{3,4,5}$\quad  Chunjie Zhang$^{1,2}$\quad Yao Zhao$^{1,2,3}$\footnotemark[2]
\\
\small{$^{1}$Institute of Information Science, Beijing Jiaotong University\quad 
$^{2}$Beijing Key Laboratory of Advanced Information Science and Network}\\
\small{$^{3}$Peng Cheng Laboratory \quad
$^{4}$Sun Yat-sen University\quad
$^{5}$MBZUAI\quad}\\
}

\maketitle

\renewcommand{\thefootnote}{\fnsymbol{footnote}}
\newcounter{somecounter}
\setcounter{somecounter}{2}
\footnotetext[1]{This work was done when Hongguang Zhu worked as an research intern in Peng Cheng Laboratory. Email: kevinlight831@gmail.com \\ \fnsymbol{somecounter} Corresponding author: yzhao@bjtu.edu.cn}
\renewcommand{\thefootnote}{\arabic{footnote}}

\begin{abstract}
Vision-Language Pretraining (VLP) has shown impressive results on diverse downstream tasks by offline training on large-scale datasets.
Regarding the growing nature of real-world data, such an offline training paradigm on ever-expanding data is unsustainable, because models lack the continual learning ability to accumulate knowledge constantly.
However, most continual learning studies are limited to uni-modal classification and existing multi-modal datasets cannot simulate continual non-stationary data stream scenarios. To support the study of Vision-Language Continual Pretraining (VLCP), we first contribute a comprehensive and unified benchmark dataset {\ourdataset} which contains over one million product image-text pairs from 9 industries. The data from each industry as an independent  task supports continual learning and conforms to the real-world long-tail nature to simulate pretraining on web data. We comprehensively study the characteristics and challenges of VLCP,  and propose a new algorithm: Compatible momentum contrast with Topology Preservation, dubbed {\ourmethod}. The compatible momentum model absorbs the knowledge of the current and previous-task models to flexibly update the modal feature. Moreover, Topology Preservation transfers the knowledge of embedding across tasks while preserving the flexibility of feature adjustment.
The experimental results demonstrate our method not only achieves superior performance compared with other baselines but also does not bring an expensive training burden. Dataset and codes are available at \url{https://github.com/KevinLight831/CTP}.
\end{abstract}
 \begin{figure}[t!]
  \centering
    \includegraphics[width=\linewidth]{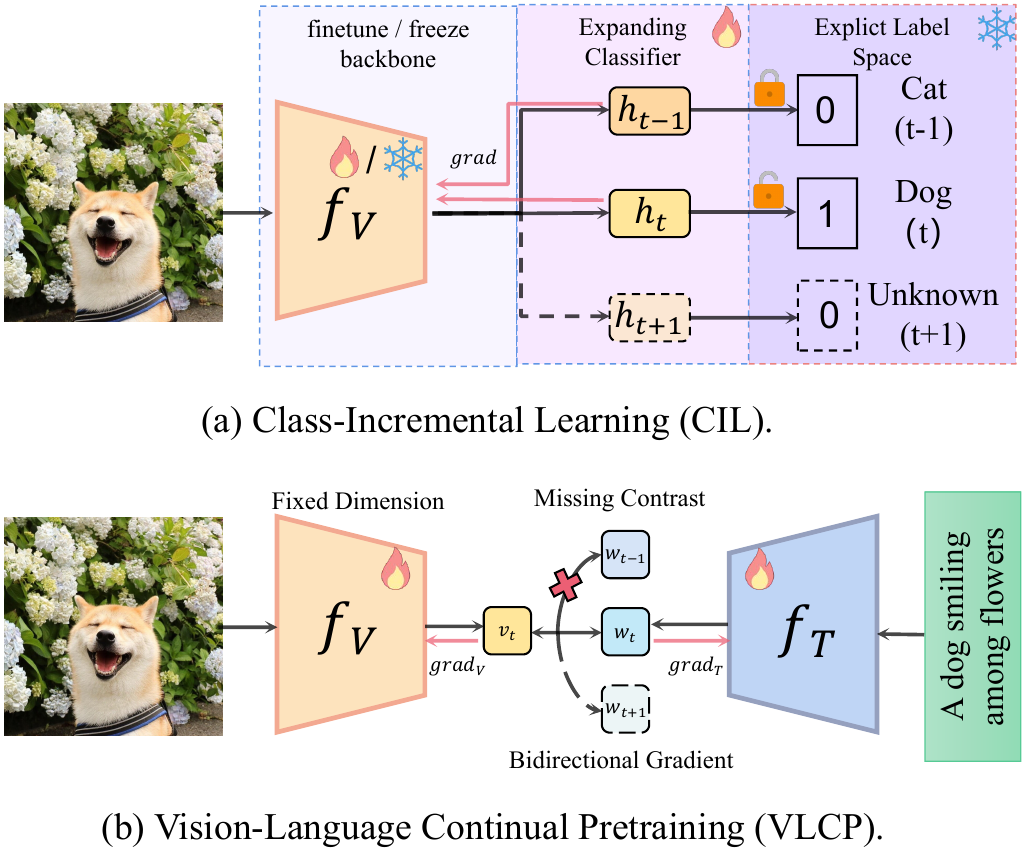}
    \vspace{-6mm}
  \caption{The traditional Class-Incremental Learning (CIL) is inflexible in the continual learning of visual concepts, which needs ever-expanding classifier parameters and endless human annotation. Moreover, it is difficult for single-class labeling to cover all visual concepts in an image. \eg, CIL only focuses on the foreground class dog and ignores the background class flower, while Vision-Language Continual Pretraining (VLCP) can flexibly represent the image content by text.
 Compared with CIL, which fixes the label space and only updates the image encoder, VLCP updates the image and text encoders simultaneously in the fixed dimension. Meanwhile, previous-task data also cannot be used as contrast samples in the continual pretraining.}
  \label{fig: intro}
  \vspace{-3mm}
\end{figure}

\vspace{-4mm}
\section{Introduction}
\vspace{-2mm}
Benefiting from the remarkable generalization ability derived from large-scale pretraining, Vision-Language Pretraining (VLP) \cite{CLIP, BLIP} has emerged as the prevalent approach for addressing downstream vision-language tasks. 
The recent advancements in artificial intelligence such as CLIP \cite{CLIP} and ChatGPT \cite{Chatgpt} have further fueled this trend of using larger models and more data.
In the long term, this computational headlong rush does not seem reasonable to move toward sustainable solutions and actually also excludes academic laboratories with limited resources. 
Current VLP paradigms all train on prepared data in advance. Nevertheless, the world is ever-changing. Offline-trained models can not evolve in a dynamic environment to continually acquire, integrate and accumulate new knowledge. Moreover, repeated offline pretraining on the ever-expanding dataset will impose growing and endless training costs. Only finetuning on new data will also suffer severe degradation due to \textit{catastrophic forgetting} \cite{mccloskey1989catastrophic}. Hence, in practical application scenarios, it is significant for VLP to continually integrate knowledge from the incoming data.

Prior studies on continual learning \cite{EWC, zhou2023deep, survey, zhang2022mining, slca} focused on supervised class-incremental learning (CIL), as \fig \ref{fig: intro}(a), which aims to maintain discriminative features for known classes and expand new classifiers to learn new classes.
However, this paradigm is inflexible due to the constant demand for laborious annotation and increasing classifier parameters.
In contrast, VLP allows for learning open-world visual patterns without explicit ``class'' concepts, which can capture more comprehensive visual concepts rather than just category-based features.
Moreover, massive web weak-aligned image-text pairs can be used as training data without extra human annotation, and no extra parameters are needed for category expansion as the output dimension is fixed for image and text encoders.

Nevertheless,  Vision-Language Continual Pretraining (VLCP) remains understudied due to the lack of datasets that satisfy both massive image-text pairs and continual tasks with discrepant knowledge.
Therefore, we contribute the first VLCP dataset \textbf{\ourdataset}, which contains more than 1 million product image-text pairs and over 3800 categories from 9 industries. 
Different task data are split according to industry (e.g. food  and clothing) to support the continual pretraining.
{\ourdataset} not only is larger than previous CIL datasets both in terms of both class number and data size. but also conforms to real-world long-tailed distributions.
As shown in the \fig \ref{fig: intro}(b), VLCP, as a new paradigm, also suffers new challenges compared with traditional CIL.
1) \textbf{Fixed-dimensional embedding:} CIL methods typically address the stability-plasticity dilemma by preserving old logits \cite{LWF} or freezing old backbone \cite{mallya2018packnet} and finetuning new classifiers. Without explicit class supervision and increasing embedding dimension, the VLCP can only adjust the fixed-dimensional shared embedding to incorporate both old and new knowledge.
2) \textbf{Missing contrast samples:}  CIL still can use the gradient from negative logits of old classes \cite{LWF, ICARL, LUCIR} even if the old data is unseen. But the lack of contrast samples from old tasks leads to suboptimal shared embedding in VLCP. 
3) \textbf{Multi-modal/task optimization:} Unlike CIL has fixed label space to optimize image encoder,  VLCP involves the complicated joint optimization of image, text, and multi-modal encoders.

Therefore, we propose a simple yet effective method, Compatible momentum contrast with Topology Preservation ({\ourmethod}), which maintains a compatible momentum model that absorbs both new and old knowledge to separately adjust uni-modal and multi-modal encoders. Moreover, different from CIL methods that distill visual features across tasks, topology preservation keeps consistent sample relationship across tasks. It not only transfers the topology knowledge of the old embedding while preserving the flexibility of feature adjustment. 
Meanwhile, to systematically investigate the vision-language continual pretraining, we extend a series of traditional CIL methods to VLCP and evaluate them in a unified setting. Interestingly, we find that the multi-modal fusion feature by masked modeling pretraining has a strong anti-forgetting ability, and the performance of continual finetuning approximates that of joint training in multi-modal retrieval. Oppositely, due to the lack of contrastive samples from different tasks, the cross-modal alignment ability suffers serious forgetting in continual pretraining and  has a big gap with joint training. Meanwhile, The experimental results show our method is not only compatible with both memory-buffer and memory-free situations, but also achieve leading performance without incurring expensive training time costs.

\vspace{1mm}\noindent\textbf{Our contributions} are as follows: 
\begin{itemize}[leftmargin=*,topsep=0pt,itemsep=0pt,noitemsep]
	\item We build the first Vision-Language Continual Pretraining (VLCP) benchmark dataset {\ourdataset} to support the study of VLCP. which contains massive image-text pairs and the continual non-stationary task data.  
	\item We systematically study the characteristics and challenges of VLCP, and establish baseline library for VLCP by extending popular continual learning methods and evaluating them in a unified setting.
	\item We propose a simple yet effective method {\ourmethod} for VLCP, which achieve both superior performance and efficient training.
\end{itemize}

\section{Related Work}
\subsection{Vision-language pretraining}
Vision-Language Pretraining (VLP) \cite{vlpsurvey} leverages large-scale web image-text pairs as pretraining data and adopts self-supervised learning (contrastive learning \cite{CLIP} or masked modeling \cite{Vl-beit}) to train the transferable image-text embeddings.
The VLP models can coarsely be divided into two paradigms according to architectures: 1) Dual-Encoder and 2) Fusion-Encoder. \textbf{Dual-Encoder} models encode images and texts respectively by separate encoders and employ cosine similarity to build the image-text alignment. The Dual-Encoder models \cite{vilbert, CLIP, ALIGN} achieve promising results on image-text retrieval with linear time complexity. However, the loose modal interaction by cosine similarity also limits the multi-modal fusion ability \cite{vqa, cornia2019show, xu2022multimodal}. Thus, the other paradigm \textbf{Fusion-Encoder} employs cross-modal attention to jointly encode images and text. The prior works \cite{UNITER, OSCAR, VinVL} use pretrained detectors to extract regional features and  Transformers \cite{transformer} for multi-modal fusion. 
However, extracting region features is computationally expensive and the joint transformer requires quadratic time complexity for retrieval tasks.
Thus, align before fuse (ALBEF) architecture \cite{ALBEF, BLIP, COCA} incorporates the image-text contrastive loss before multi-modal fusion and replaces the detector with VIT \cite{VIT} for the end-to-end pretraining.
It not only eliminates the burden of object detection pre-processing but also keeps multi-modal interaction by the top fusion layers and linear retrieval time complexity by the bottom dual encoders.

Nevertheless, current VLP models are all trained in a joint manner using prepared data and keep fixed pretrained parameters. 
In the long term, they cannot constantly accumulate knowledge and evolve themselves to accommodate the dynamic world. Thus, we concentrate on the vision-language continual pretraining based on the ALBEF architecture and evaluate the cross-modal alignment and multi-modal fusion capabilities in the continual environment. 

\subsection{Continual Learning}
Continual learning aims to overcome catastrophic forgetting and integrate novel knowledge in a sequential fashion where old data are unavailable.
Conventional continual learning methods mainly focus on image classification tasks. They can be roughly categorized into three groups: 
1) \textbf{Regularization-based methods} \cite{EWC, MAS, SI, AFEC, RWalk, LWF} limit the plasticity of the model to address catastrophic forgetting by regularizing important parameters or knowledge distillation. Although these methods alleviate forgetting to some extent without storing old samples, they cannot get satisfactory performance in some complex datasets \cite{wu2019large} and challenging settings \cite{mai2022online}. 
2) \textbf{Architecture-based methods} \cite{rusu2016progressive, serra2018overcoming, mallya2018packnet,  li2019learn} keep old parameters fixed while growing and allocating weights for learning new data. These methods can expand sub-networks or focus more on the specific part of network modules. However, these models require task identity to condition the network at test time, which is impractical for more realistic and task-agnostic settings \eg retrieval tasks. Additionally, as the number of tasks increases, the parameters of the added sub-networks become very huge, which is also not suitable for application deployment.
3) \textbf{Replay-based methods} \cite{ER, ICARL,LUCIR} apply extra memory to store a few samples from previous tasks or learn to generate pseudo data and train with the current data together. Based on this simple yet effective idea, recent methods further improve and achieve state-of-the-art performance by involving different sampling strategies. However, the memory size and the training complexity will be enlarged significantly and unaffordable as the growth of tasks, especially for costly large-scale pretraining.

However, It is an inflexible way to get the foundation model with continual learning ability through image classification. Firstly, most real-world data can not be accurately represented semantics by simple category labeling, and class annotation is labor-intensive.
Secondly. the ever-expanding classifier will also bring endless growth of parameters. 
In contrast, VLP does not require explicit "class" labeling and can cover wider visual concepts by text, while its fixed output dimensions can continually support downstream tasks without increasing parameters.

 \begin{figure*}[!t]
  \centering
    \includegraphics[width=\linewidth]{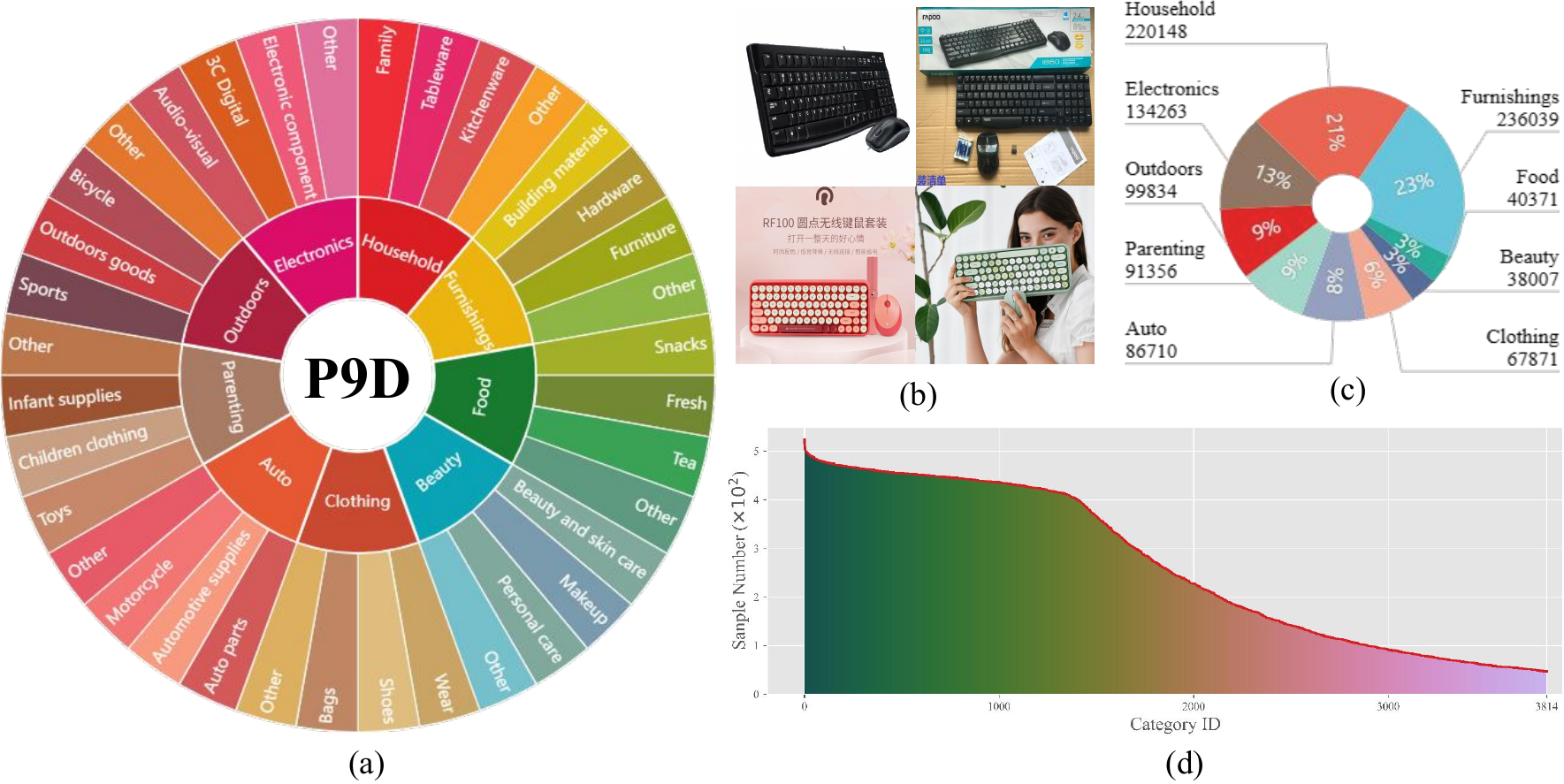}
  \caption{The dataset statistic of {\ourdataset}. (a) The abbreviated  hierarchical categories structure of {\ourdataset}. (b) Some multi-domain mixing examples about the product `Keyboard and mouse set'. (c) The quantity distribution of {\ourdataset} train set. (d) The sample number distribution of category in {\ourdataset}. The red line represents the sample number of each category in decreasing order.
  }
  \label{fig: P9D}
\end{figure*}
\section{Dataset}
Traditional CIL datasets \cite{mnist,cifar} usually use simple images with single-semantic and have the same sample number for different tasks. This ideal setting is difficult to simulate the real-world data  which is noisy and long-tail distributed. Meanwhile, despite massive web data collected by some VLP datasets \cite{CC3M,CC12M}, The simple random partitioning will make each chunk still conform to the overall distribution and unable to simulate the continual environment.
Considering the rich product samples in e-commerce websites, which not only conform to the real web-data nature but also have category, industry, and title information as weakly semantic correspondence, we use e-commerce data to construct the first vision-language continual pretraining dataset {\ourdataset} and establish the unified evaluation benchmark.
\subsection{Dataset Split}
As the \fig \ref{fig: P9D}(c), {\ourdataset} includes more than 1 million image-text pairs of real products. According to the industry name of products, {\ourdataset} is divided into 9 tasks to sequential training (default order) which are Household, Furnishings, Food, Beauty, Clothing, Auto, Parenting, Outdoor, and Electronics. We select 1,014,599 image-text pairs for training, and 2,846 pairs as the test set of cross-modal retrieval. 4,615 and 46,855 pairs as the query set and gallery set of multi-modal retrieval. The quantity distributions of the four sets are consistent and more details can refer to the appendix.
Considering the descriptions of similar products may be very similar, to avoid the situation that one image corresponds to multiple captions affecting the overall evaluation of cross-modal retrieval, we save one sample for the categories owning more than 100 samples in the training set. Meanwhile, For the query set, the number of samples per class is about 0.5\% of that in the training set.
\subsection{Data Analysis.}
\noindent {\bf Real-world Web Data}:
We collect massive commodity data from e-commerce websites and split task identity according to industry to simulate real-world rich-concept but non-stationary data streams. 
Different from the existing CIL datasets with clear images and labels, our images present the characteristics of multi-domain mixing (as \fig \ref{fig: P9D}(b)). e.g., multiple backgrounds, amorphous watermarks, occlusion, and multi-view. 
Meanwhile, although only text can be used in training, {\ourdataset} also includes product class labels to evaluate the fusion feature clustering by multi-modal retrieval.

\vspace*{2mm}\noindent {\bf Rich Categories}:
Some recent papers can obtain preferable classification results on traditional CIL datasets by finetuning \cite{l2p} and even freezing \cite{thengane2022clip} the weight of pretrained models. 
We suppose that the traditional benchmark datasets contain limited classes (e.g. MNIST \cite{mnist} and CIFAR-100 \cite{cifar}) and thus cannot adequately evaluate the continual learning methods in the face of powerful generalized pretrained models. 
Unlike these datasets with the narrow space of category labels, our dataset contains over 3800 categories and divides each task in a way that matches the real-world industry domains. Therefore, our setting is more challenging and realistic. The abbreviated  hierarchical categories structure of {\ourdataset} is shown in the \fig \ref{fig: P9D}(a), and more comparisons with other datasets can refer to the appendix.

\vspace*{2mm}\noindent {\bf Real-World Distribution}: 
The conventional CIL datasets consider a balanced distribution for each task but ignore the nature of long-tailed distributions in the real world. In contrast, our dataset contains a different number of categories for each task and the long-tailed distribution aligns well with real-world scenarios. The \fig \ref{fig: P9D}(d) shows the sample number distribution of the categories.

 \begin{figure*}[t!]
  \centering
    \includegraphics[width=\linewidth]{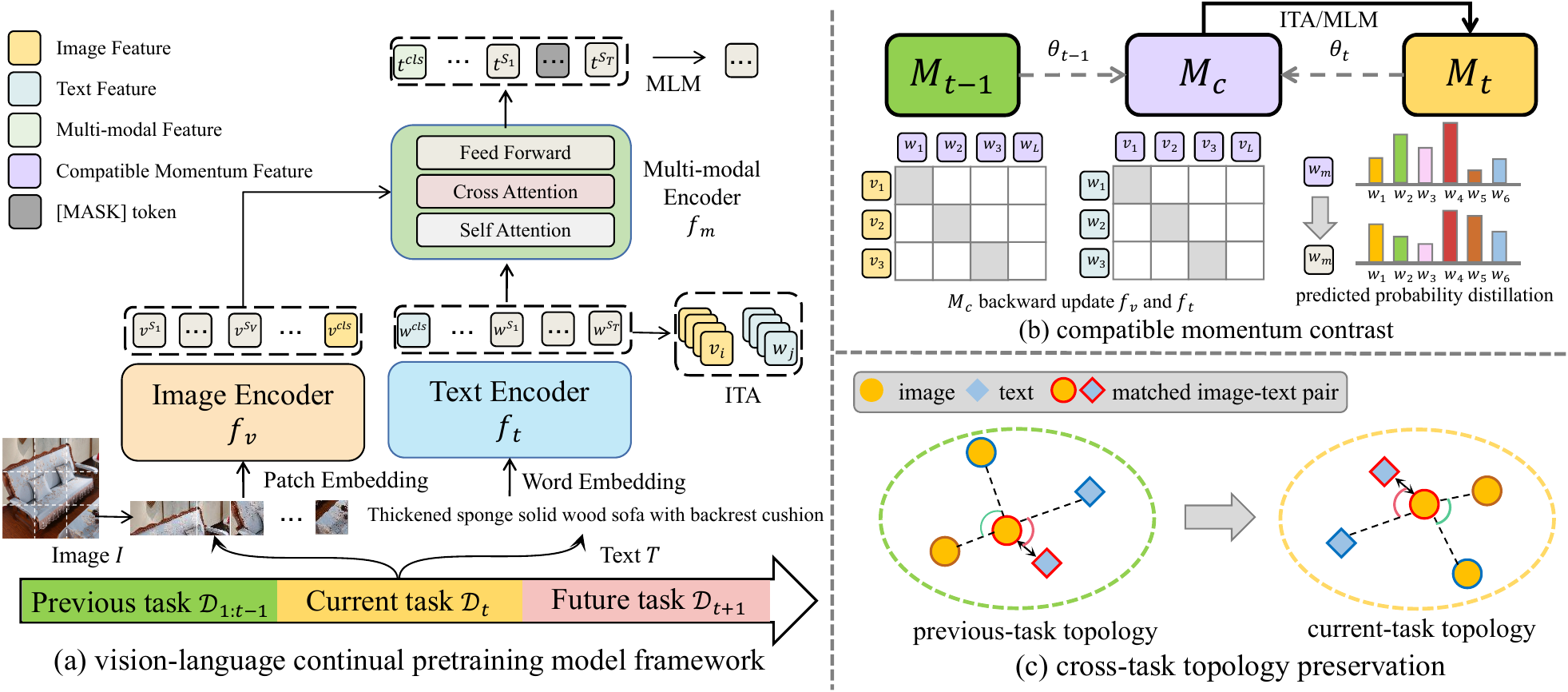}
    \vspace{-5mm}
  \caption{Illustration of the proposed vision-language continual pretraining method {\ourmethod}. (a) shows the overall continual pretraining model architecture, and (b) shows the interactive adjustment of the compatible momentum model $M_c$ and current training model $M_t$. The $M_c$ absorb the parameter of current model $M_t$ and previous-task model $M_{t-1}$, and in turn constrains the update of $M_t$. (c) shows that the topology relationship of previous task is maintained as much as possible while allowing the overall embedding to be updated.}
  \label{fig: main}
  \vspace{-4mm}
\end{figure*}
\vspace{-2mm}
\section{Methodology}
\subsection{Preliminary} 
\noindent {\bf Problem Setting}: 
We propose a  vision-language continual pretraining (VLCP) setting, where models are supposed to be sequentially trained on $T$ tasks data $\mathcal{D} = \{\mathcal{D}_1, \mathcal{D}_2,..., \mathcal{D}_T\}$. In the $t$-th task, the whole sub-dataset $\mathcal{D}_t = \{(I_i^t, T_i^t)\}_{i=1}^{N_t}$ contains the $N_t$ image-text pairs where $I_i^t$ and $T_i^t$ respectively denote the $i$-th image and the corresponding text description of the $t$-th task. 
Because the old task data is unseen in the continual setting, the VLP model trained on the current dataset $\mathcal{D}_t$ needs to resist forgetting and performs well for all learned datasets $\{\mathcal{D}_1, \mathcal{D}_2,..., \mathcal{D}_{t}\}$. 

\vspace*{2mm}\noindent {\bf Model Architecture}: 
As shown in \fig \ref{fig: main}, We use a 12-layer ViT-B/16 \cite{VIT} as the image encoder $f_v$, and initialize it with weights pretrained on ImageNet-1k from \cite{DEIT}. The image $I$ is encoded into patch feature sequence $f_v(I)=\{{v}^\mathrm{cls},{v}^1,...,{v}^N\}$. Meanwhile, We use the first and the last 6 layers of BERT$_\mathrm{base}$ \cite{BERT} model to initialize the text encoder $f_t$ and multi-modal encoder $f_m$. The text T is encoded into word feature sequence $f_t(T)=\{{w}^\mathrm{cls},{w}^1,...,{w}^N\}$. Then, the text features $f_t(T)$ will fuse with the image features $f_v(I)$ through cross attention at each layer of  multi-modal encoder $f_m$. Two linear transformations $g_v$ and $g_t$ are map the ${v}^\mathrm{cls}$ and ${w}^\mathrm{cls}$ to the low-dimensional (256-d) representations for Image-Text Alignment (ITA). It can be denoted $v = g_v({v}^\mathrm{cls})$ and $w = g_w({w}^\mathrm{cls})$. 

Given a pair of image-text data $(I_i, T_j)$ in the current $D_t$, the image-text similarity function $s(I_i, T_j)$ is defined as the cosine similarity $s(I_i, T_j) = \frac{v_i^Tw_j}{\|v_i\|\|w_j\|}$.
Given a batch of $B$ pairs, the model uses the symmetric cross-entropy loss over the $B \times B$ similarity matrix to optimize the parameters.
The image-to-text loss $\mathcal{L}_{i2t}$ and the text-to-image loss $\mathcal{L}_{t2i}$ are formulated as:
\vspace{-2mm}
\begin{equation}
\begin{aligned}
\mathcal{L}_{i2t} &= -\frac{1}{B} \sum_{i=1}^{B}\log \frac{\exp(s(I_i, T_i)/\tau)}{\sum_{j=1}^{B} \exp(s(I_i, T_j)/\tau)},\\
\mathcal{L}_{t2i} &= -\frac{1}{B} \sum_{i=1}^{B}\log \frac{\exp(s(T_i, I_i)/\tau)}{\sum_{j=1}^{B} \exp(s(T_i, I_j)/\tau)},
\end{aligned}
\label{eq:2}
\end{equation}
where $\tau$ is the temperature parameter. The Image-Text Alignment (ITA) loss is defined as : $\mathcal{L}_{ita} = \frac{\mathcal{L}_{i2t} + \mathcal{L}_{t2i}}{2}$.

In Masked Language Modeling (MLM), give an image-text pair, we randomly mask out the words with probability of 15\% \cite{BERT,ALBEF}, and replace masked ones ${w}_\mathrm{m}$ with the special token \texttt{[MASK]}. 
The goal is to predict these masked tokens based on their surrounding words and the image features, by minimizing the cross-entropy loss:
\begin{equation}
\mathcal{L}_{mlm} = -\mathbb{E}_{(I,\hat{T}) \sim D_t} \mathrm{H} (y^\textrm{m}, p^\textrm{m}_{\theta}(I,\hat{T})),
\label{eq:3}
\end{equation}
where the $y^\textrm{m}$ is the one-hot vocabulary distribution where the masked token has a probability of 1, and $p^\textrm{m}_{\theta}(I,\hat{T})$ is the predicted probability of model $\theta$ for masked token ${w}_\mathrm{m}$.
The total VLP loss is defined as $\mathcal{L}_{VLP} = \mathcal{L}_{ita} + \mathcal{L}_{mlm}$.

\subsection{Compatible Momentum Contrast} \label{subsec:CMU}
Due to the fixed-dimensional embedding, VLP cannot isolate old and new knowledge like CIL by extending projection parameters. Therefore, both the vision and language embeddings need to be constantly adjusted to simultaneously accommodate the old and new image-text pairs. 
In order to review old knowledge and adapt to the new task, we use the momentum model $M_{c}$ initialized by the previous-task trained model $M_{t-1}$ as additional supervision of the current training model $M_{t}$.
Some single-modal \cite{MOCO} or vison-language \cite{wenlan} pretraining works also adopt momentum model as a temporal ensembling method \cite{frenchself} to smoothly guide the training.
However, the traditional momentum model is updated by only parameters of training model. With the accumulation of training steps, it will be gradually affected by the new model and also suffers catastrophic forgetting. Moreover, recklessly maintaining old knowledge will also make the model lose the plasticity to acquire new knowledge. 
Therefore, we propose the compatible momentum update which synchronously absorbs the old and new knowledge:
\begin{equation}
\theta_{c} \leftarrow m \cdot \theta_{c} + \frac{1-m}{2} \cdot \theta_{t-1} + \frac{1-m}{2} \cdot \theta_{t},
\label{eq:5}
\end{equation}
where $m \in [0,1)$ is the momentum coefficient and $\theta$ is the model parameters.
The adjustment of models is interactive. Compatible momentum model $M_c$ updates parameters $\theta_{c}$ through the previous-task model $M_{t-1}$ and training model $M_{t}$. In turn, the training model $M_{t}$ are optimized by the back-propagation of momentum contrast and affects the parameters $\theta_{t}$ to be passed in the next step. To further steadily update uni-modal encoders $f_v$ and $f_t$, we maintain two dynamic queues to preserve the $K$ negative image/text features. The image features $v_i^{c}$ and text feature $w_i^{c}$ from compatible momentum encoders are constantly pushed into visual queue $Q^I=\{v_1^{c},v_2^{c}, \cdots, v_{N_Q}^{c}\}$ and text queue $Q^T=\{w_1^{c},w_2^{c}, \cdots, w_{N_Q}^{c}\}$ which $N_Q = K+B$. The compatible momentum contrastive losses about the vision and language encoders can be formulated as follows:
\begin{equation}
\begin{aligned}
\mathcal{L}_{i2t}^c &= -\frac{1}{B} \sum_{i=1}^{B}\log \frac{\exp(s(I_i, Q^T_i)/\tau)}{\sum_{j=1}^{N_Q} \exp(s(I_i, Q^T_j))/\tau)},\\
\mathcal{L}_{t2i}^c &= -\frac{1}{B} \sum_{i=1}^{B}\log \frac{\exp(s(T_i, Q^I_i)/\tau)}{\sum_{j=1}^{N_Q} \exp(s(T_i, Q^I_j)/\tau)}, \\
\label{eq:6}
\end{aligned}
\vspace{-5mm}
\end{equation}

Similarly, $\mathcal{L}_{ita}^c = \frac{\mathcal{L}_{i2t}^c + \mathcal{L}_{t2i}^c}{2}$.
$\mathcal{L}_{ita}^c$ constraints that the contrastive relation of image-text pairs is still workable between the encoders of the compatible momentum model and current model. It allows slow adjustment of uni-modal encoders. Besides, The compatible momentum model also provides the soft predicted probability for masked language modeling loss.
\begin{equation}
\begin{aligned}
\mathcal{L}_{mlm}^c = -\mathbb{E}_{(I,\hat{T}) \sim D_t} \mathrm{H} (p^{\textrm{m}}_{\theta_{c}}(I,\hat{T}), p^{\textrm{m}}_{\theta_t}(I,\hat{T})),
\label{eq:7}
\end{aligned}
\end{equation}
Thus, Compatible Momentum Contrastive loss can be defined as $\mathcal{L}_{CMC} = \mathcal{L}_{ita}^c + \mathcal{L}_{mlm}^c$.

\subsection{Topology Preservation} \label{subsec:TP}
Although the compatible momentum contrast can flexibly adjust the output feature of uni-modal and fusion encoder, It does not directly transfer relationship knowledge of  samples across tasks and the model may forget the overall topology of prior embedding to obtain sub-optimal performance. Unlike CIL can receive the gradient of old classes according to labels, VLCP has no gradient from old contrast samples and the image-text encoder is bidirectional synchronization adjustment according to the image-text similarity. To integrate the old and new knowledge while maintaining the topology relationship of prior tasks, we constrained the mini-batch sample relationships of the current and previous-task models to be consistent.
Specifically, it is mainly divided into cross-modal topology preservation loss $\mathcal{L}_{c}$ and same-modal topology preservation loss $\mathcal{L}_{s}$.
To image $I = \{I_1,I_2,...,I_B\}$ and text $T=\{T_1,T_2,...,T_B\}$, we define the image-to-text similarity distribution as follows: 
\begin{equation}
\mathcal{P}^{i2t} = \frac{\exp(s(I, T_i)/\tau)}{\sum_{i=1}^{B} \exp(s(I, T_i))/\tau)},
\label{eq:9}
\end{equation}
and the definition of text-to-image similarity distribution $\mathcal{P}^{t2i}$ is similar. Thus, the $\mathcal{L}_{c}$ can be formulated as the cross-entropy $H$ between current model $M_t$ and previous-task model $M_{t-1}$:
\begin{equation}
\small
\mathcal{L}_{c}=\frac{1}{2}\mathbb{E}_{(I,T) \sim D_t} [H(\mathcal{P}^{i2t}_{\theta_{t-1}}, \mathcal{P}^{i2t}_{\theta_{t}})+H(\mathcal{P}^{t2i}_{\theta_{t-1}}, \mathcal{P}^{t2i}_{\theta_{t}})],
\label{eq:10}
\end{equation}

In the same-modal topology preservation, the similarity of the same sample is 1 under both old and new models. However, such a large similarity would suppress the relational distillation of other unmatched pairs \cite{DKD}. Thus, we conduct the simple strategy that changes the similarity at the diagonal of ${s}(I, I)$ from 1 to a minimum (\eg -1000) to exclude the ``apical dominance'' of matched samples. We formulate the changed matrix as $\hat{s}(I, I)$ and the $\mathcal{L}_{s}$ can be formulated as:
\begin{equation}
\hat{\mathcal{P}}^{i2i} = \frac{\exp(\hat{s}(I, I_i)/\tau)}{\sum_{i=1}^{B} \exp(\hat{s}(I, I_i))/\tau)},
\label{eq:11}
\end{equation}
\begin{equation}
\small
\mathcal{L}_{s}=\frac{1}{2}\mathbb{E}_{(I,T) \sim D_t} [H(\hat{\mathcal{P}}^{i2i}_{\theta_{t-1}}, \hat{\mathcal{P}}^{i2i}_{\theta_{t}})+H(\hat{\mathcal{P}}^{t2t}_{\theta_{t-1}}, \hat{\mathcal{P}}^{t2t}_{\theta_{t}})],
\label{eq:12}
\end{equation}
Thus, the overall continual pretraining loss is as follows:
\begin{equation}
\mathcal{L} = \mathcal{L}_{VLP} + \mathcal{L}_{CMC} + \mathcal{L}_{c} + \mathcal{L}_{s},
\label{eq:13}
\end{equation}

\begin{table*}[t!]
\centering
\setlength{\tabcolsep}{4pt}
\small
{
\begin{tabular}{lc ccc|ccc|c||ccc}
\toprule[1pt]
 \multirow{2}*{\bfseries Methods} &\multirow{2}*{\bfseries \makecell{Training\\hours}}& \multicolumn{7}{c||}{ Cross-modal Retrieval}  &\multicolumn{3}{c}{ Multi-modal Retrieval} \\ 
\cmidrule(lr){3-9} 
\cmidrule(lr){10-12}
& &{TR@1} & {TR@5} & {TR@10} & {IR@1} & {IR@5} & {IR@10} & {Rm} & {mAP@1} & {mAP@5} & {mAP@10}\\%
\midrule
JointT  &-- &61.31  &87.17   &91.67  &61.60  &86.79 &91.95  &80.08  &63.79  &70.10  &67.40   \\
\midrule
\multicolumn{8}{l}{\bfseries Memory-Free} \\
\midrule
SeqF    &3.4 &34.79  &62.83  &72.03   &35.73  &63.63  &72.14  &56.86  &62.15  &68.03  &65.30 \\
SI \cite{SI}             &4.9 &34.96  &63.60  &73.01   &35.84  &63.49  &73.44  &57.39  &61.32  &67.48  &64.78 \\
MAS \cite{MAS}             &4.1 &36.65  &64.69  &73.93   &37.25  &64.55  &74.39  &58.57  &62.62  &68.63  &65.79 \\
EWC \cite{EWC}            &4.6 &37.28  &65.11  &74.38   &38.05  &65.57  &75.05  &59.24  &62.99  &68.75  &65.62 \\
AFEC \cite{AFEC}           &8.7 &37.67  &66.34  &74.67   &38.79  &66.48  &75.16  &59.85  &62.58  &68.44  &65.66 \\
LWF \cite{LWF}            &4.0 &37.63  &66.55  &75.09   &38.26  &66.62  &75.26  &59.90  &62.34  &68.64  &65.94 \\
RWalk \cite{RWalk}          &6.2 &37.77  &68.10  &76.70   &38.83  &67.39  &76.74  &60.92  &62.41  &68.43  &65.59 \\
\rowcolor{danred}Our:\ourmethod      &4.0 &\textbf{43.43}  &\textbf{72.10}  &\textbf{80.08}  &\textbf{43.39}  &\textbf{71.15}  &\textbf{79.06} &\textbf{64.87}  &62.64 &68.21 &65.10\\
\midrule
\multicolumn{8}{l}{\bfseries Memory-Buffer} \\
\midrule
MoF \cite{ICARL}           &4.8 &42.87  &71.93  &80.60   &43.71  &72.03  &80.67  &65.30  &62.84  &68.79  &65.98 \\
LUCIR \cite{LUCIR}          &5.5 &43.82  &73.68  &82.04   &44.38  &73.54  &80.99  &66.41  &61.06  &67.57  &64.74 \\
ER  \cite{ER}            &4.4 &45.19  &73.40  &81.97   &44.97  &72.70  &81.10  &66.56  &62.21  &68.39  &65.80 \\
Kmeans \cite{ER}         &4.7 &46.17  &74.65  &82.33   &45.92  &73.68  &81.80  &67.26  &62.77  &68.82  &66.04 \\
ICARL \cite{ICARL}          &5.3 &45.85  &74.63  &82.57   &46.24  &73.23  &81.83  &67.39  &63.51  &69.20  &66.39 \\
\rowcolor{danred}Our:\ourmethod$+$ER &4.7 &\textbf{50.53} &\textbf{77.62}  &\textbf{84.57}  &\textbf{49.79}  &\textbf{76.77}  &\textbf{84.47}  &\textbf{70.63}  &62.62  &68.68  &65.81 \\
\bottomrule[1pt]
\end{tabular}
\vspace{-2mm}
\caption{The final cross-modal and multi-modal retrieval performance comparison with different Memory-Free and Memory-Buffer continual learning baselines.}
\label{tab: result}
}
\vspace{-3mm}
\end{table*}

\begin{figure*}[t!]
  \centering
    \includegraphics[width=1\linewidth]{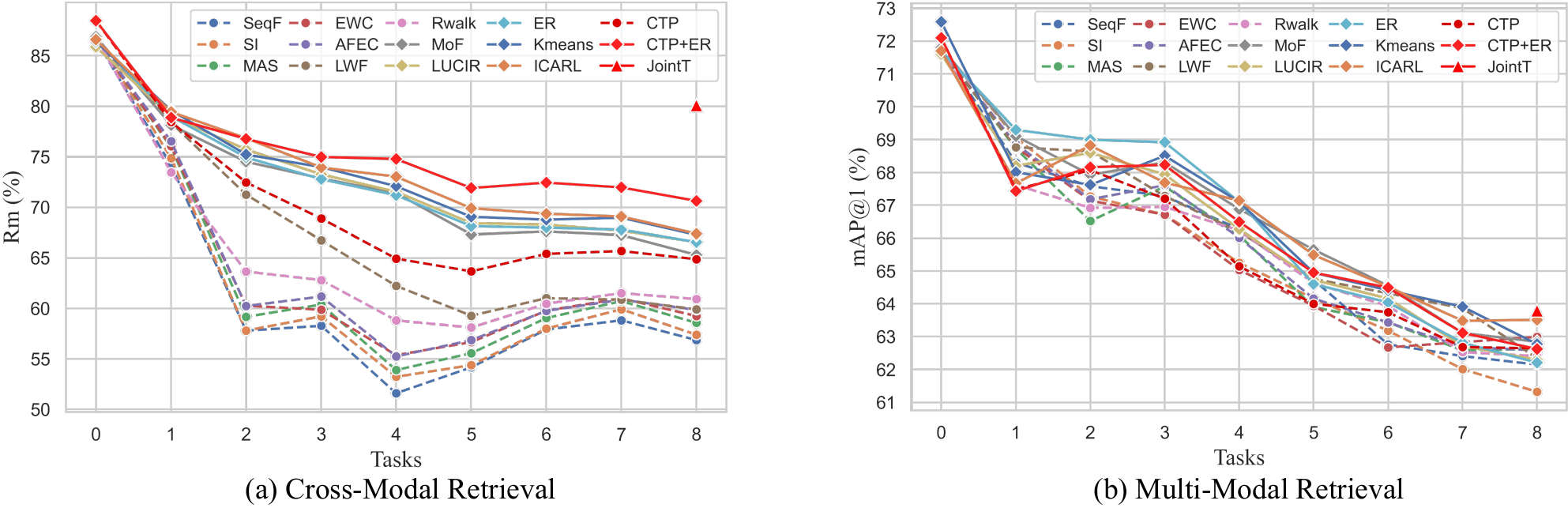}
    \vspace{-6mm}
  \caption{The performance curve of different methods on the continual learning. For Task $i$, the model $\theta_i$ only test on the merged test set of learned tasks $\{\mathcal{D}_0,\mathcal{D}_1,...,\mathcal{D}_i\}$.
  }. 
  \label{fig: line}
  \vspace{-6mm}
\end{figure*}
\section{Experiment}
\subsection{Experimental Setup}
\vspace*{0mm}\noindent {\bf Implementation Details}
Regarding offline pretrained VLP models have established widespread generalization, direct finetuning may involve knowledge leakage and weaken the accurate evaluation of continual learning ability. Therefore, we do not load the weight of the pretrained VLP model for initialization.
For each task, All models are trained for 5 epochs on 4 NVIDIA A100 GPUs with batch size 128 per GPU.
We use the AdamW \cite{AdamW} optimizer with a weight decay of 0.05, and the learning rate is set as 1e-4 and decays to 1e-6 following a cosine schedule. 
We take random image crops of resolution 224 $\times$ 224 during pretraining and also apply RandAugment. 
The maximum sequence length of tokens is limited to 30. The momentum parameter is set as 0.9, and the queue size K is set as 1024.
The cross-modal alignment temperature $\tau$ = 0.07.

\vspace*{2mm}\noindent {\bf Evaluation Setting.}
With the knowledge accumulation in continual learning, the evaluation galleries of cross-modal and multi-modal retrieval also need to expand by merging the data of new tasks.
Similar to the standard vision-language modeling setting\cite{CLIP, ALBEF, BLIP}, For cross-modal retrieval, we measure the performance with Recall at K (R@K, K=1,5,10) \cite{vse++}, which is defined as the proportion of ground truth being retrieved at top-K of the ranking list. Also, we use Rm as the overall metric, which is defined as the mean of R@K of both text and image retrieval.

The multi-modal encoder can fuse the multi-modal information and produce a comprehensive representation to predict  masked words. 
Thus, we use the multi-modal output \texttt{[CLS]} token to retrieve samples of the same class to evaluate the zero-sample clustering ability of multi-modal fusion features for similar products. 
It will not introduce extra interference from finetuning on downstream tasks. (\eg classification needs to train new classification head from scratch).
Because each query here corresponds to multiple targets with the same class, multi-modal retrieval considers both the number and ranking of targets in the top-K retrieved candidates. 
For multi-modal retrieval, we adopt mean Average Precision (mAP@N ) \cite{weyand2020google, zhan2021product1m, dong2022m5product} as the evaluation metrics. mAP@N is computed as follows:
\begin{equation}
\begin{aligned}
\textrm{mAP@N} &= \frac{1}{Q} \sum_{i=1}^{Q} \frac{1}{\mathbf{m_q}} \sum_{k=1}^{N} P_q(k) \delta_q(k) \\
P_q(k) &=\frac{R_q(k)+1}{k}
\end{aligned}
\end{equation}
where $Q$ is the total number of multi-modal queries and $m_q$ is the total target number of the $q$-th query in the retrieved top-N relevant candidates from gallery. $P_q(k)$ is the precision at rank $k$ for the $q$-th query, and the $\delta_q(k)$ is a binary indicator function that returns 1 when the $k$-th prediction is correct for the $q$-th query and 0 otherwise. $R_q(k)$ is the current sum of returned correct predictions at the rank $k$. 

\vspace*{2mm}\noindent {\bf Comparison Methods.}
To verify the effectiveness of our method, we compare it with several popular continual learning methods. Since all competitors are originally proposed for classification settings. We replace the instance-label cross-entropy loss with image-text contrastive loss to train the VLP model and add the Masked Language Modeling loss to train the multi-modal fusion capability as follows \cite{ALBEF}. Besides, We use joint training (JointT) of all seen samples as the upper-bound performance and sequential finetuning (SeqF) as the lower-bound. 
All baselines can be separated into Memory-Free and Memory-Buffer methods based on the availability of old samples.
For the former. we evaluate representative regularization-based methods such as EWC \cite{EWC}, SI \cite{SI}, MAS \cite{MAS}, AFEC \cite{AFEC}, RWalk \cite{RWalk}, and LwF \cite{LWF}. As for the latter, we evaluate different replay sampling strategies: ER \cite{ER}, Mean-of-Feature (MoF)\cite{ICARL}, and Kmeans \cite{ER}, and some replay-based methods: ICARL\cite{ICARL} and LUCIR \cite{LUCIR}. 
We must emphasize that although Memory-Buffer methods usually have higher performance than Memory-Free methods in CIL, they bring expensive storage costs in large-scale pretraining, and performance is directly related to the replay sample size. For the comprehensive and unified study of VLCP, to all Memory-Buffer methods, we maintain a fixed-size buffer of 10000 image-text pairs (about 1\% dataset) and continually update the stored samples.
The detailed introduction for each method can refer to the appendix.

\vspace{-1mm}
\subsection{Experimental Results}
\noindent {\bf Cross-modal \vs Multi-modal.}
From the \fig \ref{fig: line},  we found an interesting phenomenon that SeqF (lower-bound) and JointT (upper bound) have a large gap (23.22\% in Rm) in cross-modal retrieval, but a small gap (1.65\% in mAP@1) in multi-modal retrieval. This phenomenon shows that multi-modal fusion is stronger anti-forgetting ability than cross-modal alignment in VLCP.
We suppose that on the one hand, The redundancy and complementarity of multi-modal information help multi-modal fusion resist the forgetting of class attributes.
On the other hand, the pretext task MLM beforehand creates a word prediction classifier that corresponds to each word of the dictionary and keeps consistency across tasks. This predefined and well-initialized label space \cite{objects} maybe train models more stably than ever-expanding label space.

\vspace*{2mm}\noindent {\bf Parameter \vs Topology Preservation.}
From the comparison of Memory-Free methods in \tab \ref{tab: result}, we observe that the regularization methods \cite{SI, MAS, AFEC, RWalk} represented by EWC \cite{EWC} perform poorly. Probably because such methods conservatively trust the old model parameters and cannot flexibly update representation to better accommodate new knowledge. Meanwhile, They introduce the extra post-processing of calculating the fisher matrix which reduces training efficiency. In contrast, benefiting from flexible updating of compatible momentum contrast and soft constraints of topology preservation, Our {\ourmethod} achieves 3.95\% improvement over the most competitive method on Rm while not bring more time costs and extra memory burden. 

\vspace*{2mm}\noindent {\bf Memory-Free \vs Memory-Buffer.}
It can be noted that the Memory-Buffer methods exhibit superior final performance and smaller performance fluctuations compared to the Memory-Free methods. This is attributed to the use of old data from the memory buffer as contrast samples, which plays the role of joint optimization in continual pretraining.
Meanwhile, we found different replay sampling strategies have similar performance, but Kmeans\cite{ER} has a slight advantage, It may be because Kmeans can unsupervised cluster features to sample representative points of current embedding without category prior. However, the Memory-Buffer methods also bring extra storage and time cost causing by old data preservation and retraining process. The results show {\ourmethod} can be further improved by 5.76\% with ER sampling strategy and outperform the second-place method by 3.24\% on Rm.

\begin{table}[!t]
\setlength{\tabcolsep}{1mm}{
\centering
\begin{tabular}{cccc|cccc} 
\toprule
  $\mathcal{L}_{CMC}$   & $\mathcal{L}_c$ & $\mathcal{L}_s$ & ER &TR@1 &IR@1 &Rm & mAP@1  \\ 
\midrule
\xmark &\xmark &\xmark &\xmark &34.79 &35.73 & 56.86  & 62.15    \\
\cmark &\xmark &\xmark &\xmark &37.81 &37.63 & 59.86  & 62.10    \\%
\xmark &\cmark &\xmark &\xmark &41.45 &40.39 & 62.50  & 61.32    \\%
\cmark &\cmark &\xmark &\xmark &42.69 &42.41 &64.38   &61.89    \\
\cmark &\xmark &\cmark &\xmark &41.92 &42.02 &63.17   &61.11    \\
\cmark &\cmark &\cmark &\xmark &43.43 &43.39 &64.87  & 62.64    \\%
\cmark &\cmark &\cmark &\cmark &50.53 &49.79 &70.63  & 62.62    \\%
\bottomrule
\end{tabular}
\vspace{-3mm}
\caption{The ablation study on each component of {\ourmethod}. the $\mathcal{L}_{CMC}$ represent the compatible momentum contrast. }.
\label{tab: ablation}
\vspace{-3mm}
}
\end{table}

\begin{table}[!t]
{
\centering
\begin{tabular}{l|cccc} 
\toprule
   Method   &TR@1 &IR@1 &Rm & mAP   \\ 
\midrule
 only  $\theta^{t-1}$                    &41.95 &42.23  &63.57  &62.06    \\ %
 only $\theta^{t}$                       &40.41 &40.34  &62.32   &62.32  \\ %
 w/o $Q$                                 &42.69 &40.83  & 63.94  & 61.63    \\
 w/o $\hat{\mathcal{P}}^{i2i}$ \& $\hat{\mathcal{P}}^{t2t}$  &40.76 &40.86  & 63.12  & 62.04    \\
\hline
 Our:\ourmethod &43.43 &43.39 &64.87  & 62.64    \\ 
\bottomrule
\end{tabular}
\vspace{-3mm}
\caption{The more detailed ablation results of {\ourmethod}.}
\vspace{-3mm}
\label{tab: more_ablation}
}
\end{table}

\subsection{Ablations}
We perform ablation of each module in the {\ourmethod} method and find that each module effectively improves the cross-modal retrieval performance from the \tab \ref{tab: ablation}. 

\vspace*{2mm}\noindent {\bf CMC \vs TP.}
It seems that cross-modal topology preservation plays a more direct anti-forgetting role than compatible momentum contrast in cross-modal retrieval, which improves 5.64\% on Rm but also decreases 0.82\% on mAP@1. When they are combined, the cross-modal and multi-modal retrieval performance were all improved by 1.88\% and 0.57\%. In addition, The same-modal topology preservation further improves by 0.49\% and 0.75\%.

\vspace*{2mm}\noindent {\bf Compatible Update \vs Single-way Update.}
As shown in \tab \ref{tab: more_ablation}, we compare compatible momentum update with the single-way momentum update from the current model $\theta^t$ and the previous-step model $\theta^{t-1}$, and find the single-way update can not get well results due to totally relay on the current or old model. 
In addition, the momentum update from $\theta^{t-1}$ is slightly higher than that from $\theta^t$ in Rm. 
This may be because the model accumulates large knowledge in continual learning, and the non-globally optimal update will lead to more forgetting in the later tasks. Therefore, it should pay more attention to the maintenance of old knowledge in the later periods of continual learning.

\vspace*{2mm}\noindent {\bf Momentum Queue.}:
We compare with the situation without the momentum queue `w/o $Q$' in \tab \ref{tab: more_ablation} and find that the queue can slightly improve the performance. We suppose the queue stores negative samples of the current task, which plays a role of smoothing training and anti-forgetting to a certain extent. 

\vspace*{2mm}\noindent {\bf Suppress Same-modal Maximum Similarity.}:
 We find it results in a negative effect and leads to performance degradation if not suppressing same-modal maximum similarity. It indicates that the maximum similarity will obscure the sample relationship and affect the performance.

\section{Conclusion}
In this paper, we build the first Vision-Language Continual Pretraining benchmark dataset {\ourdataset} which contains over 1 million image-text pairs and task data with discrepant knowledge to simulate the continual pretraining environment. Further, we comprehensively study new characteristics and challenges of VLCP, and propose a new approach {\ourmethod} which combines the compatible momentum contrast and topology preservation to flexibly update model to accommodate the ever-changing embedding. It can achieve the superior performance while ensuring efficient training.   

{\small
\bibliographystyle{ieee_fullname}
\bibliography{egbib}
}

\newpage

\appendix
\begin{center}
\large \textbf{Appendix}
\end{center}

\noindent 
This supplementary document mainly provides more information about our {\ourdataset} dataset and implementation details of the baseline methods. 
Besides, we provide the pseudocode of {\ourmethod} and more experimental studies. 

\section{{\ourdataset} Dataset.}
\subsection{Dataset Split.}
\begin{figure}[t]
  \centering
    \includegraphics[width=\linewidth]{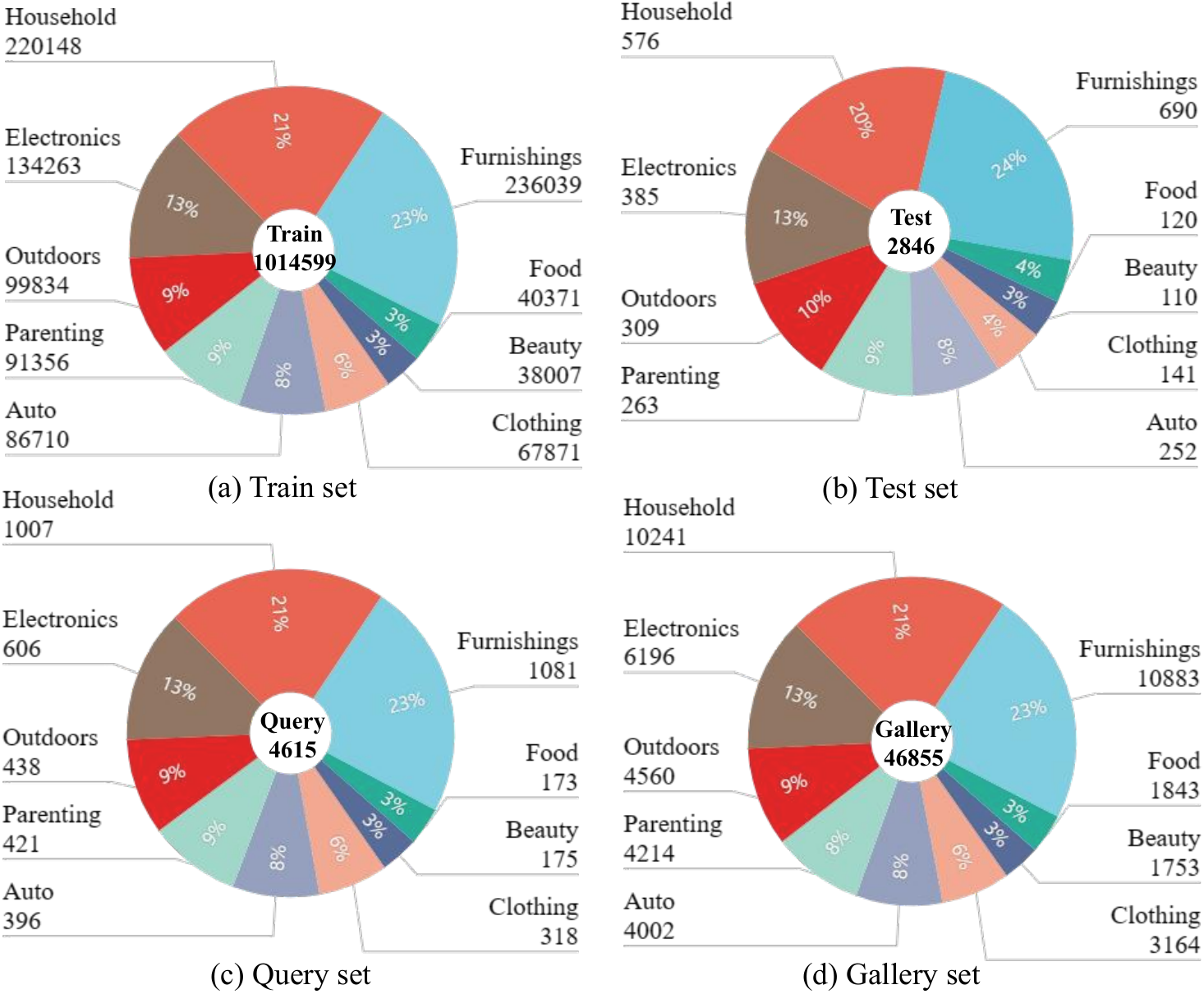}
    \vspace{-5mm}
  \caption{The quantity distribution of different task data is consistent for the four subsets of our {\ourdataset} dataset. }
  \label{fig: fourset}
  \vspace{-4mm}
\end{figure}

\fig \ref{fig: fourset} shows the quantity distribution of each subset of our {\ourdataset}. The different subsets have a consistent quantity distribution across tasks, and this consistent distribution ensures the comprehensive and unified evaluation for pretraining. Different from the training set, the test set (cross-modal retrieval evaluation) and query set (multi-modal retrieval evaluation) need to be further filtered by humans. The filter criterion is that the text describes the image content as accurately as possible while ensuring that the test/query set is proportional to the training set for the same category.
\subsection{Image-Text Examples.}
\begin{figure*}[t!]
  \centering
  \includegraphics[width=0.9\linewidth]{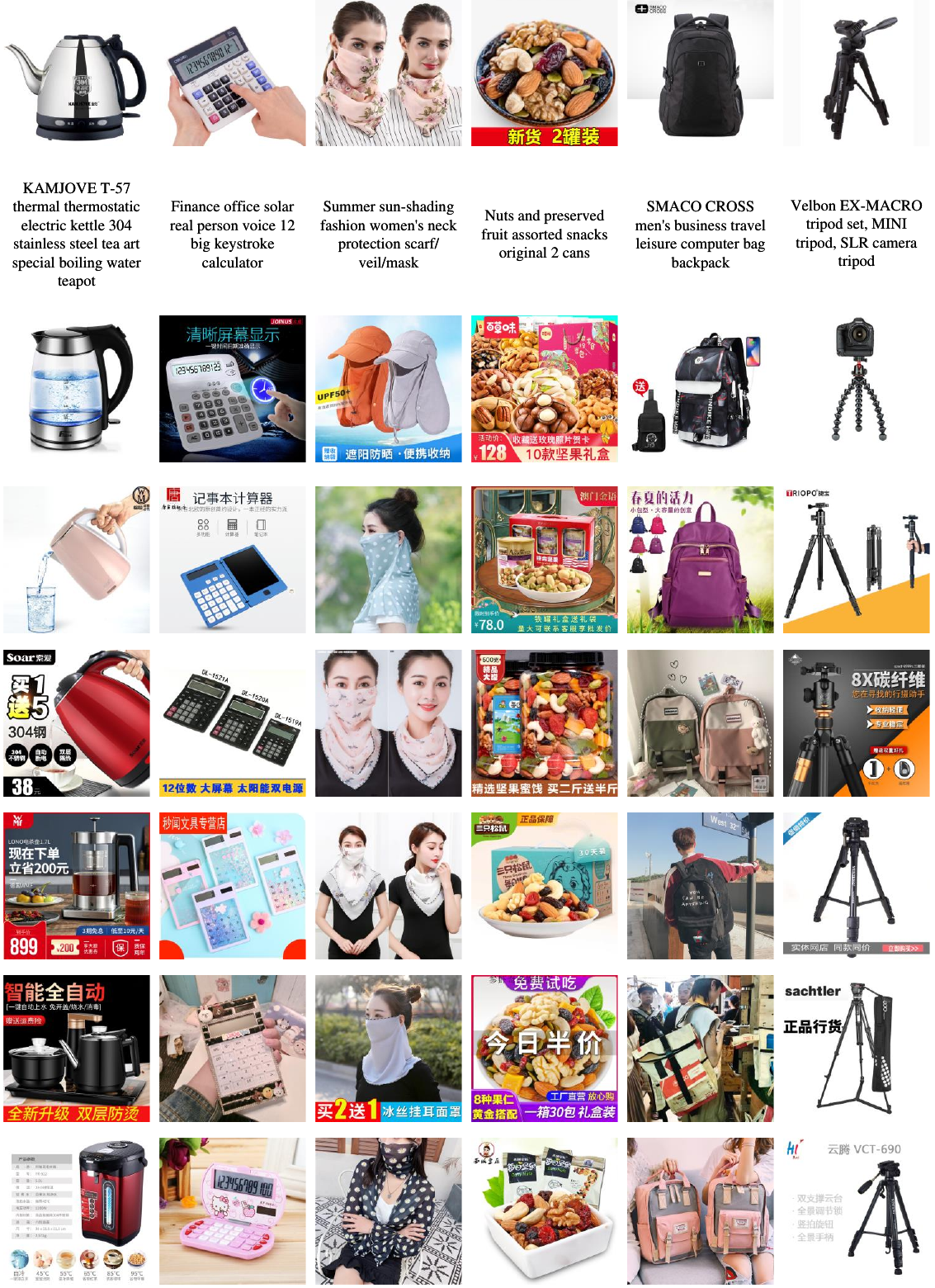}
  \caption{Some examples of our dataset. The first and second rows are the corresponding image-text pair. For simplicity, the rest rows show the images from the same class.}
  \label{fig: sample}
\end{figure*}

\fig \ref{fig: sample} shows some image-text examples. We show some images of same class and keep one described caption for simplicity. It shows that real-world web data is noisy and multi-domain mixing. There are prevalent and complicated situations in the web image domain, such as complex backgrounds, amorphous watermarks, irrelevant objects, and occlusion.

\section{Details of Baseline Methods.}
Because these baseline methods are originally proposed for continual learning on the image classification task. Thus, we re-implement them to adapt the setting of vision-language pretraining. In addition to the replacement of the main optimization loss, we present the implementation details of each comparison method as follows:
\subsection{Memory-Free methods}
\vspace*{2mm}\noindent {\bf EWC \cite{EWC}} is the classical regularization methods. It maintains the old model parameters $\theta_{t-1}$ and an important matrix $\Omega$ with the same scale as the model. EWC builds an additional regularization loss to remember the old parameters according to the important matrix. Because the model $\theta_{t-1}$ at the last task stores the old knowledge, consolidating important parameters can fix the knowledge from being forgotten. The training loss can be formulated as:
\begin{equation}
\mathcal{L}_{EWC} = \mathcal{L}_{VLP} + \frac{1}{2}\lambda\Sigma_k \Omega_k(\theta_{t,k}-\theta_{t-1,k})^{2}, 
\label{eqn: ewc}
\end{equation}
where the $\theta_{t-1,k}$ denotes the $k$-th parameter after training last task data $\mathcal{D}_{t-1}$. $\Omega_k$ means the important weight of the $k$-th parameter and is calculated by the Fisher Information Matrix (FIM) in the EWC method.

\vspace*{2mm}\noindent {\bf SI \cite{SI}} considers that the EWC is conducted at the end of each task and will ignore the optimization dynamics over the entire training trajectory. Thus, SI online estimates the importance weight $\Omega_k$ by its contribution (backward gradient) to the total loss variation. However, this online strategy need to backpass the gradient twice for each iteration. In the re-implement, we store the gradient of each parameter by retaining the forward graph.  

\vspace*{2mm}\noindent {\bf MAS \cite{MAS}} calculate $\Omega_k$ by a unsupervised way. Specifically, It accumulates important measures based on the sensitivity of predictive results (output features) to parameter changes. In our re-implement, we sum the norm of the visual, textual, and multi-modal features as the predictive result to calculate the importance.

\vspace*{2mm}\noindent {\bf RWalk \cite{RWalk}} combines the regularization terms of SI \cite{SI} and EWC \cite{EWC} to integrate their advantages. In each iteration, Rwalk simultaneously consolidates the parameter by considering the online importance weight from SI method and the offline important weight from EWC method.

\vspace*{2mm}\noindent {\bf AFEC \cite{AFEC}} proposes to actively forget the old knowledge that interferes with the learning of new tasks for continual learning. Specifically, It introduces the extra forward-step trained model $\theta^{\star}_t$ as the expansion and collaboratively guides the update of the current model with the EWC method. Similar to EWC, the training loss can be formulated as:
\begin{equation}
\begin{aligned}
 \mathcal{L_{AFEC}} &= \mathcal{L}_{VLP} + \frac{1}{2}\lambda\Sigma_k \Omega_k(\theta_{t,k}-\theta_{t-1,k})^{2} \\
&+ \frac{1}{2}\lambda_e\Sigma_k \Omega_k^{\star}(\theta_{t,k}-\theta_{t,k}^{\star})^{2}, 
\end{aligned}
\label{eqn: ewc}
\end{equation}
where $\theta^{\star}_t$ is the parameter of forward-step trained model and $\lambda_e$ is the FIM of $\theta^{\star}_t$. 

\vspace*{2mm}\noindent {\bf LWF \cite{LWF}} aligns the representations of previous-step and current models for all new arriving data. We maintain one reference model whose parameters are copied from the previous-step trained model and align the image and text representation of the reference and current model by the cross-entropy loss for each iteration. 

\subsection{Memory-Buffer methods}
For the memory updating processing, the replay buffer will delete some old samples and add some new samples according to the size of the new task data.

\vspace*{2mm}\noindent {\bf ER \cite{ER}} is a popular sample selection strategy. It uses the reservoir sampling \cite{JeffreyScottVitter1985RandomSW} randomly stores a fixed number of training samples for each input batch and each sample has the same probability of being replaced.  

\vspace*{2mm}\noindent {\bf Kmeans \cite{ER}} use the Kmeans clustering to process all samples of the current task and set the number of clusters to the number of corresponding replaced samples. Then the cluster-center samples are chosen to update the buffer.

\vspace*{2mm}\noindent {\bf MoF \cite{ICARL}} is first proposed by ICARL \cite{ICARL} and selects samples that are closest to the feature mean of each class. Because vision-language pretraining has no class concept, we choose the samples that are closest to the multi-modal feature mean of the current task.

\vspace*{2mm}\noindent {\bf ICARL \cite{ICARL}} perform knowledge distillation on both buffer samples and new samples. The sample selection strategy is Mean-of-Feature (MoF). In our implementation, we combine the LWF term to optimize the current model. 

\vspace*{2mm}\noindent {\bf LUCIR \cite{LUCIR}} proposes to utilize a cosine classifier to avoid the influence of the biased classifier and encourage similar feature orientation of the new and previous-step models. In our implementation, we replace the regular projected linear layer with the cosine normalizing linear layer. In addition, we constrain that the similarity of same-modal embedding from new and old models is big as possible. However, the inter-class distance constraint of the original paper \cite{LUCIR} cannot be re-implemented because there is no class label in vision-language pretraining.

\begin{table*}[!t]
\setlength{\tabcolsep}{3mm}{
\centering
\begin{tabular}{l|cc|ccc|c} 
\toprule
  Dataset   & Train Samples & Categories & Modal &Objects &Continual Task Split &URL \\ 
\midrule
\multicolumn{5}{l}{\bfseries Popular Class-Incremental Learning Dataset} \\
\midrule
Oxford Flowers \cite{Oxford-Flowers} &2,040 &102  &image &single  &yes &\href{https://www.robots.ox.ac.uk/~vgg/data/flowers/102/}{Link}\\
VOC Actions \cite{VOC-Actions} &3,102 &11  &image &single  &yes &\href{http://pascallin.ecs.soton.ac.uk/challenges/VOC/voc2012/}{Link}\\ 
MIT Scenes \cite{MIT-Scenes} &5,360 &67  &image &single  &yes &\href{http://web.mit.edu/torralba/www/indoor.html}{Link}\\ 
CUB200-2011 \cite{CUB200-2011} &5,994 &200  &image &single  &yes &\href{http://www.vision.caltech.edu/datasets/cub_200_2011/}{Link}\\
FGVC-Aircraft \cite{FGVC-Aircraft} &6,666 &100  &image &single  &yes &\href{https://www.robots.ox.ac.uk/~vgg/data/fgvc-aircraft/}{Link}\\
Letters \cite{Letters}  &6,850 &52  &image &single  &yes &\href{https://archive.ics.uci.edu/ml/datasets/Letter+Recognition}{Link}\\
Stanford Cars \cite{Stanford-Cars} &8,144 &196  &image &single  &yes &\href{https://ai.stanford.edu/~jkrause/cars/car_dataset.html}{Link}\\
SVHN \cite{SVHN}  &73,257 &10  &image &single  &yes &\href{http://ufldl.stanford.edu/housenumbers/}{Link} \\
CIFAR10 \cite{cifar} &50,000 &10  &image &single &yes  &\href{https://www.cs.toronto.edu/~kriz/cifar.html}{Link} \\
CIFAR100 \cite{cifar} &50,000 &100  &image &single &yes  &\href{https://www.cs.toronto.edu/~kriz/cifar.html}{Link} \\
MNIST \cite{mnist} &60,000 &10  &image &single &yes  &\href{http://yann.lecun.com/exdb/mnist/}{Link}\\
Tiny-ImageNet \cite{CLsurvey} &80,000 &200 &image &single &yes  &\href{https://www.kaggle.com/c/tiny-imagenet}{Link}\\
ImageNet-100 \cite{ImageNet-100} &130,000 &100  &image &single &yes &\href{https://openaccess.thecvf.com/content_CVPR_2019/html/Hou_Learning_a_Unified_Classifier_Incrementally_via_Rebalancing_CVPR_2019_paper.html}{Link} \\
CORe50 \cite{core50} &120,000 &50  &image &single  &yes &\href{https://vlomonaco.github.io/core50/}{Link}\\
\midrule
\multicolumn{5}{l}{\bfseries Popular Multi-Modal Dataset} \\
\midrule
Flickr30K \cite{flickr} &29,000 &--  &image-text &multi &no &\href{http://nlp.cs.illinois.edu/}{Link} \\
COCO \cite{COCO} &113,287 &80  &image-text &multi &no &\href{https://cocodataset.org/#home}{Link} \\
Visual Genome  \cite{Visual-Genome} &108K &-- &image-text &multi &no &\href{https://visualgenome.org/}{Link}\\
FashionGen \cite{FashionGen} &325,536 &-- &image-text &multi &no &\href{https://fashion-gen.com/}{Link} \\
SBU \cite{SBU} &875K &-- &image-text &multi &no &\href{http://www.cs.virginia.edu/~vicente/sbucaptions/}{Link}\\
GQA \cite{GQA} &1M &-- &image-text &multi &no &\href{https://cs.stanford.edu/people/dorarad/gqa/}{Link}\\
VQA v2.0 \cite{VQA-v2} &1.1M &-- &image-text &multi &no &\href{https://visualqa.org/}{Link} \\
CC3M \cite{CC3M} &3.1M &--  &image-text &multi  &no &\href{https://github.com/google-research-datasets/conceptual-captions}{Link}\\
CC12M \cite{CC12M} &12M &--  &image-text &multi  &no &\href{https://github.com/google-research-datasets/conceptual-12m}{Link}\\
YFCC-100M \cite{YFCC-100M} &100M &-- &image-text  &multi &no &\href{http://projects.dfki.uni-kl.de/yfcc100m/}{Link}\\
LAION-400M \cite{LAION-400M} &400M &-- &image-text &multi &no &\href{https://laion.ai/laion-400-open-dataset/}{Link}\\
\midrule
\textbf{Our: {\ourdataset}} &1,014,599 &3,814 &image-text &multi &yes &--\\
\bottomrule
\end{tabular}
\caption{The overview of datasets about continual learning and vision-language pretraining domains. `Categories' means the number of classes in the corresponding dataset and `--' means not mentioned. `Objects' means the number of labeled/described objects in images. `Continual Task Split' means the dataset contains different data chunks with discrepant semantic concepts and supports to simulate the continual environment. `URL' means the hyperlink of corresponding dataset websites.}
\label{tab:dataset_comparsion}
}
\end{table*}

\section{Dataset Comparison.}
In \tab \ref{tab:dataset_comparsion}, we present the comparison of our {\ourdataset} with popular datasets from the continual learning domain  \cite{CLsurvey} and multi-modal domain \cite{pengchengVLP}.
We observe that traditional continual learning datasets have a small number of data samples with limited classes (mostly at the thousand level), and only for single-target class labeling without detailed text description. In addition, although existing multi-modal datasets contain a large number of web image-text pairs, their data are too noisy and mixed to conform to the data split for continual tasks. In contrast, our {\ourdataset} contains abundant image-text pairs to support vision-language pretraining. Besides, Each task contains rich semantic concepts, and different generalized semantic domains.  It can support the simulation of continual learning environments.

\subsection{Class-Incremental Learning Datasets}
\noindent {\bf Oxford Flowers \cite{Oxford-Flowers}, MIT Scenes \cite{MIT-Scenes}, CUB200-2011 \cite{CUB200-2011}, Stanford Cars \cite{Stanford-Cars}, FGVC-Aircraft \cite{FGVC-Aircraft}, VOC Actions \cite{VOC-Actions}, Letters \cite{Letters}, SVHN \cite{SVHN}.}
Aljundi \etal \cite{expert-gate, MAS} propose to use a sequence of 8 highly diverse recognition tasks as continual tasks. This sequence is composed of 8 different topics, going from flowers, scenes, birds, and cars, to aircrafts, actions, letters, and digits.

\vspace*{2mm}\noindent {\bf CIFAR10/100 \cite{cifar}} consists of 60,000 32 $\times$ 32 color images in 10 classes, with 6,000 images per class. There are 50,000 training images and 10,000 test images.
The CIFAR100 dataset has 100 classes containing 600 images each. There are 500 training images and 100 testing images per class and the 100 classes can be grouped into 20 superclasses.

\vspace*{2mm}\noindent {\bf MNIST \cite{mnist}} is a large handwritten digits dataset. It has 60,000 samples as the training set and 10,000 samples as the test set.

\vspace*{2mm}\noindent {\bf Tiny-ImageNet \cite{CLsurvey}} first used in the study of continual learning by Matthi \etal \cite{CLsurvey}. This is a subset of 200 classes from ImageNet \cite{JiaDeng2009ImageNet} and the image size is rescaled to 64 $\times$ 64. Each class contains 500 samples subdivided into training (80\%) and validation (10\%), and 50 samples for evaluation. 

\vspace*{2mm}\noindent {\bf ImageNet-100 (SubImageNet) \cite{ImageNet-100}} is a 100-class random sample subset of ImageNet. It contains 130,000 images for training and 5,000 images for testing.

\vspace*{2mm}\noindent {\bf CORe50 \cite{core50}} is a collection of 50 objects collected in 11 distinct domains, where 8 of them (120,000 samples) are used for training, and the rest are used as a single test set (45,000).

\subsection{Multi-modal Datasets}
\noindent {\bf Flickr30K \cite{flickr}} is obtained by extending the corpus of Hodosh \etal \cite{MicahHodosh2013FramingID} and the image topic contain everyday scenes and activities.
There are 31,783 images associated with five manually annotated captions each, and 29,000 images are used for training.

\vspace*{2mm}\noindent {\bf COCO \cite{COCO}} is built based on MSCOCO dataset \cite{COCO}.
It consists of 123,287 images and each image is annotated with 5 captions. There are 113,287 training images, 5000 test images, and 5000 validation images. COCO and Flickr30K datasets are often used as the retrieval evaluation dataset for large-scale vision-language pretraining.

\vspace*{2mm}\noindent {\bf Visual Genome \cite{Visual-Genome}}
is proposed to help to develop of visual understanding tasks (\ie image caption and visual question answering, \etc ) by mining the relationships between objects.  The dataset contains more than 108K images and each image has about 35 objects, 26 attributes, and 21 pairwise relationships.

\vspace*{2mm}\noindent {\bf FashionGen \cite{FashionGen}} contains 325,536 1360$\times$1360 fashion images and each image has a paragraph-length caption as the description. Six different angles are photographed for all fashion items.

\vspace*{2mm}\noindent {\bf SBU \cite{SBU}} is collected and filtered from Flickr.com. It is usually used as the subset of vision-language pretraining \cite{ALBEF, BLIP, UNITER}.  

\vspace*{2mm}\noindent {\bf GQA \cite{GQA}} is a balanced dataset with 1.7M samples which is mainly proposed for visual reasoning and compositional question answering. 

\vspace*{2mm}\noindent {\bf VQA v2.0 \cite{VQA-v2}} is proposed to reduce the language biases that existed in previous VQA datasets. It consists of around 1.1M image-question pairs and 13M corresponding answers based on 200K MSCOCO images.

\vspace*{2mm}\noindent {\bf CC3M \cite{CC3M}} is a dataset annotated with conceptual captions and the image-text samples are mainly collected from the web. It contains about 3.3M image-description pairs.

\begin{algorithm*}[t]
\caption{Pseudocode of {\ourmethod} in a PyTorch-like style.}
\label{alg:code}
\algcomment{\fontsize{7.2pt}{0em}\selectfont \textbf{CE}: cross entropy loss; \textbf{eye\_like}: create an identity matrix with the same size as input.
}
\definecolor{codeblue}{RGB}{106, 155, 88} 
\lstset{
  backgroundcolor=\color{white},
  basicstyle=\bf\fontsize{7.2pt}{7.2pt}\ttfamily\selectfont,
  columns=fullflexible,
  breaklines=true,
  captionpos=b,
  commentstyle=\fontsize{7.2pt}{7.2pt}\color{codeblue},
  keywordstyle=\fontsize{7.2pt}{7.2pt},
}
\vspace{-2mm}
\begin{lstlisting}[language=python]
# F, M, R: training, momentum, and reference (previous-task) model
# m, t, q_v, q_t: momentum, temperature, visual and textual queues

M.params, R.params = F.params, F.params  # initialize momentum and reference model
for (image, text) in loader:  # load a minibatch with N image-text pairs
    feat_v, feat_t = F.get_featuere(image, text)
    multimodal_out = F.multimodal_fusion(image, text, text.mask)
    S_i2t, S_t2i = feat_v @ feat_t.T, feat_t @ feat_v.T # cross-modal similarity
    S_i2i, S_t2t = feat_v @ feat_v.T, feat_t @ feat_t.T # same-modal similarity
    
    ita_loss = CE(S_i2t/t,eye_like(S_i2t))+CE(S_t2i/t,eye_like(S_t2i)) # image-text contrastive loss
    mlm_loss = CE(multimodal_out, labels=text.labels, mask=text.mask) # mask language modeling loss
    loss = ita_loss/2 + mlm_loss # conventional loss of the current task

    # compatible momentum update
    M.params = m*M.params+(1-m)/2*F.params+(1-m)/2*R.params 
    feat_vm, feat_tm = M.get_featuere(image, text) 
    multimodal_out_m = M.multimodal_fusion(image, text, text.mask)
    enqueue(q_v, feat_vm.detach(), q_t, feat_tm.detach()) # enqueue current features
    S_i2t_m, S_t2i_m = feat_v @ q_t.T, feat_t @ q_v.T

    # compatible momentum contrast
    ita_loss_m = CE(S_i2t_m/t, eye_like(S_i2t_m)) + CE(S_t2i_m/t, eye_like(S_t2i_m))
    mlm_loss_m = CE(multimodal_out,labels=multimodal_out_m.logits,mask=text.mask)
    loss += ita_loss_m/2 + mlm_loss_m
    dequeue(q_v, q_t) # dequeue earliest features

    # topology preservation
    feat_vr, feat_tr = R.get_featuere(image, text) 
    S_i2t_r, S_t2i_r = feat_vr @ feat_tr.T, feat_tr @ feat_vr.T
    S_i2i_r, S_t2t_r = feat_vr @ feat_vr.T, feat_tr @ feat_tr.T
    loss_sm = CE(S_i2i/t,S_i2i_r/t,mask=eye_like(S_i2i))+ CE(S_t2t/t,S_t2t_r/t,mask=eye_like(S_t2t))
    loss_cm = CE(S_i2t/t, S_i2t_r/t) + CE(S_t2i/t, S_t2i_r/t)
    loss += loss_sm + loss_cm
    
    # parameter update
    loss.backward()
    update(F.params)
\end{lstlisting}\vspace{-3mm}
\end{algorithm*}


\vspace*{2mm}\noindent {\bf CC12M \cite{CC12M}} is a product of the urgent need for large-scale data with rapidly developing vision-language pre-training. The authors of CC3M relax the image-text filters and obtain the larger dataset CC12M.

\vspace*{2mm}\noindent {\bf YFCC-100M \cite{YFCC-100M}} totally contains 100 million media objects (99.2 million photos, 0.8 million videos) collected from Flickr.com.

\vspace*{2mm}\noindent {\bf LAION-400M \cite{LAION-400M}} is filtered using pre-trained CLIP \cite{CLIP} and contains 400 million image-text pairs.

\section{Algorithm}
The \alg \ref{alg:code} shows the training pipeline of our {\ourmethod} in the task data $\mathcal{D}_t \in \{\mathcal{D}_1, \mathcal{D}_2,..., \mathcal{D}_T\}$.

\section{More Experiments.}
\subsection{Momentum Setting.}
For the first task ($t=0$), There is no previous-step model and the current model has not adapted to the product domain. Thus, the momentum $m$ of the first task is set to 0.995, and we keep it the same for all ablation studies. Because we find that the training loss oscillates and fails to converge if $m$ is 0.9 in the first task. 

For the following tasks, we set $m$ as 0.9 and we also do the parameter-sensitive study about $m$. The results of \tab \ref{tab: momentum} show the training on the $\{1,2,.., T\}$ task is not sensitive to the setting of compatible momentum $m$. We suspect this is due to the fact that the model accepts parameters from both the previous-step and current models and is less prone to biased updates. In addition to the main vision-language pretraining loss, the compatible momentum contrastive loss is an auxiliary loss for continual learning. Thus, the model is more robust to momentum parameter selection and does not easily collapse \cite{MOCO}.

 \begin{table}[!t]
\setlength{\tabcolsep}{3mm}
{
\centering
\begin{tabular}{l|cccc} 
\toprule
Method   &TR@1 &IR@1 &Rm & mAP   \\ 
\midrule
only $\theta^{t-1}$                    &41.95 &42.23  &63.57  &62.06    \\ %
only $\theta^{t}$                       &40.41 &40.34  &62.32   &62.32  \\ %
m=0.7             &43.64 &43.04  &64.83  &62.95  \\
m=0.8             &43.68 &43.11  &64.74  &62.36    \\
m=0.9             &43.43 &43.39  &64.87  &62.64    \\ 
m=0.99            &43.50 &43.01  &64.75  &62.23    \\
m=0.995           &44.27 &42.23  &65.04  &61.63    \\
\bottomrule
\end{tabular}
\caption{The results of momentum selection experiment.}
\label{tab: momentum}
}
\end{table}

\begin{table*}[t!]
\centering
\setlength{\tabcolsep}{4pt}
\small
{
\begin{tabular}{lccc|ccc|c||ccc}
\toprule[1pt]
 \multirow{2}*{\bfseries Methods} & \multicolumn{7}{c||}{ Cross-modal Retrieval}  &\multicolumn{3}{c}{ Multi-modal Retrieval} \\ 
\cmidrule(lr){2-8} 
\cmidrule(lr){9-11}
&{TR@1} & {TR@5} & {TR@10} & {IR@1} & {IR@5} & {IR@10} & {Rm} & {mAP@1} & {mAP@5} & {mAP@10}\\%
\midrule
JointT                       &60.72  &86.05   &91.74  &61.98  &86.82 &91.85  &79.86  &64.07  &70.09  &67.33   \\
\midrule
\multicolumn{8}{l}{\bfseries Memory-Free} \\
\midrule
SeqF                         &37.81  &64.69  &74.00   &38.05  &64.23  &74.46  &58.87  &61.86  &68.08  &65.18 \\
SI \cite{SI}                 &38.51  &64.41  &74.84   &38.90  &65.14  &74.14  &59.32  &61.56  &67.63  &64.69 \\
MAS \cite{MAS}               &39.81  &66.97  &75.86   &40.86  &66.87  &75.93  &61.05  &61.65  &67.78  &65.23 \\
EWC \cite{EWC}               &39.11  &67.96  &77.09   &41.46  &68.73  &77.30  &61.94  &62.17  &68.11  &65.22 \\
AFEC \cite{AFEC}             &40.13  &68.17  &77.62   &41.57  &68.69  &76.99  &62.19  &61.60  &67.66  &64.96 \\
LWF \cite{LWF}               &41.18  &67.29  &76.39   &39.81  &67.81  &76.32  &61.31  &61.76  &68.12  &65.20 \\
RWalk \cite{RWalk}           &39.04  &67.85  &78.00   &40.20  &68.94  &77.65  &61.95  &62.40  &68.43  &65.60 \\
\rowcolor{danred}Our:\ourmethod  &\textbf{45.96}  &\textbf{73.47}  &\textbf{80.85}  &\textbf{44.98}  &\textbf{72.34}  &\textbf{80.25} &\textbf{66.31}  &61.08 &67.20 &64.19\\
\midrule
\multicolumn{8}{l}{\bfseries Memory-Buffer} \\
\midrule
MoF \cite{ICARL}             &43.92  &72.28  &80.99   &45.01  &73.19  &81.03  &66.07  &61.37  &67.65  &64.79 \\
LUCIR \cite{LUCIR}           &45.36  &72.91  &80.92   &45.61  &73.68  &80.74  &66.54  &61.89  &67.92  &65.31 \\
ER  \cite{ER}                &44.59  &72.87  &81.20   &45.92  &73.05  &80.89  &66.42  &62.32  &68.35  &65.42 \\
Kmeans \cite{ER}             &45.19  &74.42  &81.83   &46.03  &73.05  &80.67  &66.53  &62.73  &68.35  &65.39 \\
ICARL \cite{ICARL}           &47.33  &75.65  &83.63   &47.61  &75.90  &83.24  &68.89  &62.54  &68.54  &65.87 \\
\rowcolor{danred}Our:\ourmethod$+$ER &\textbf{51.05} &\textbf{79.20}  &\textbf{86.23}  &\textbf{51.30}  &\textbf{78.60}  &\textbf{85.45}  &\textbf{71.97}  &61.58  &67.65  &64.78 \\
\bottomrule[1pt]
\end{tabular}
\vspace{-2mm}
\caption{The final cross-modal and multi-modal retrieval performance comparison when conducting the reversed task order.}
\label{tab: reversed}
}
\vspace{-3mm}
\end{table*}

\subsection{Reverse Task Order.}
In the main text, all experiments are conducted in the default task order. To study the impact of task order on the performance ranking, we supplement a check experiment with the reversed task order \footnote{Electronics, Outdoor, Parenting, Auto, Clothing, Beauty, Food, Furnishings, and Household}. The \tab \ref{tab: reversed} shows the results of all baselines and our method in the reversed task order.

The result shows that although there are some changes in the ranking of some methods with similar performance, the overall performance ranking is still consistent with the performance ranking of default task order. Additionally, our method {\ourmethod} exhibits superior performance in both continual learning scenarios (Memory-Free and Memory-Buffer), even when the order of tasks is changed. It indicates that the task order can affect the performance value of final result but not the performance ranking of our method. Our method consistently outperforms in different task order settings.

\section{License}
Our {\ourdataset} dataset is released under CC BY-NC-SA 4.0 license and can freely be used for non-commercial purposes.
The collection of data has obtained permission from the relevant websites. Once a conflict of interest, our group reserves all the rights for the final explanation.

\end{document}